# Application of Gist SVM in Cancer Detection

**Mrs S. Aruna,** Dept of Computer Applications
**Dr S. P. Rajagopalan,** Prof Emeritus
**Mr L. V. Nandakishore,** Department of Mathematics
**Dr M.G.R University Chennai-95, India**

**ABSTRACT:** In this paper, we study the application of GIST SVM in disease prediction (detection of cancer). Pattern classification problems can be effectively solved by Support vector machines. Here we propose a classifier which can differentiate patients having benign and malignant cancer cells. To improve the accuracy of classification, we propose to determine the optimal size of the training set and perform feature selection. To find the optimal size of the training set, different sizes of training sets are experimented and the one with highest classification rate is selected. The optimal features are selected through their F-Scores.

**Introduction**

Pattern classification problems can be effectively solved by Support Vector Machines (SVMs) [CV95; Vap98]. SVMs find applications in data mining, bioinformatics, computer vision, and pattern recognition. There is a need to accelerate SVM training as the size of the training data sets is becoming larger.

In this paper, we study the application of SVM in the prediction of cancer using the GIST SVM. SVM is a class of learning methods that can be used for classification. The problem of the optimal feature selection plays a vital role in SVM which ensures high performance classification. Non-optimal features ensure robustness since SVMs are trained for maximized margins.

**39**

This paper is structured as follows. Section 1 gives a brief introduction to SVM classifier and the f-score method used for feature selection, In Section 2, we discuss the results obtained and concluding remarks are given in Section 3 to address further research issues.

## 1. Materials and Methods

### 1.1. Support Vector Machines

Support vector machines are a set of related supervised learning methods used for classification and regression. Linear classification when used for classification, the SVM algorithm creates a hyper plane that separates the data into two classes with the maximum-margin. Fig. 1 shows the SVM classifier with hyper plane. Given training examples labeled either "yes" or "no", a maximum-margin hyper plane is identified which splits the "yes" from the "no" training examples, such that the distance between the hyper plane and the closest examples (the margin) is maximized.

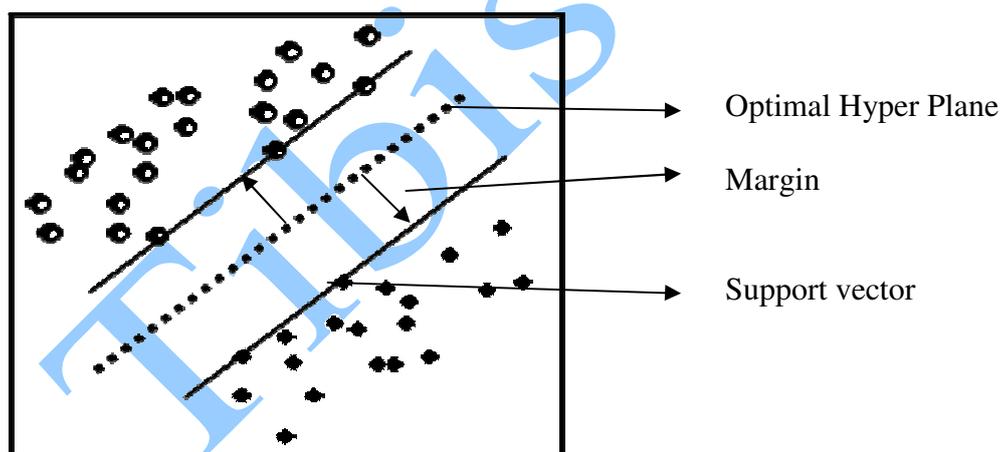

**Figure 1. SVM classifier**

The use of the maximum-margin hyper plane is motivated by Vapnik Chervonenkis theory, which provides a probabilistic test error bound that is minimized when the margin is maximized.

The parameters of the maximum-margin hyper plane are derived by solving a quadratic programming (QP) optimization problem. There exist several specialized algorithms for quickly solving the QP problem that arises from SVMs.

**40**

In training SVMs, the decision boundaries are determined directly from the training data so that the generalization ability is maximized. Therefore, the generalization ability of the SVM is quite different from those of other classifiers, especially when the number of training data is small. Here we propose to use Gist (http://svm.sdsc.edu) to perform all the training and testing. Gist is an implementation of the SVM algorithm.

To apply SVM in training and testing the practical data sets, one has to determine the size of the training set and the optimal features. We attempt to find the optimum size for the training set and the test set because the time complexity of training an SVM is in the order of $n^2$, where $n$ is the number of training samples [Pla98; Joa98].

We used Wisconsin breast cancer (WBC) dataset which has 699 instances (Benign: 458 Malignant: 241) of which 16 instances has missing attribute values removing that we have 683 instances of which 444 benign and 239 are malignant. This breast cancer database was obtained from the University of Wisconsin Hospitals, Madison from Dr. William H.Wolberg available in UCI Machine learning depository http://archive.ics.uci.edu/ml).

Features are computed from a digitized image of a Fine Needle Aspiration (FNA). The feature extraction process is performed as follows: An FNA is taken from the breast mass. This material is then mounted on a microscope slide and stained to highlight the cellular nuclei. A portion of the slide in which the cells are well differentiated is then scanned using a digital camera and a frame-grabber board and identified nine visually assessed characteristics of an FNA sample, which he considered relevant to diagnosis. The resulting data set is well-known as the Wisconsin Breast Cancer Data.

Most breast cancers are detected by the patient as a lump in the breast. The majority of breast lumps are benign so it is the physician's responsibility to diagnose breast cancer, that is, to distinguish benign lumps from malignant ones. There are three available methods for diagnosing breast cancer: Mammography, FNA (with visual interpretation) and surgical biopsy. The reported sensitivity (i.e., ability to correctly diagnose cancer when the disease is present) of Mammography varies from 68% to 79%, of FNA with visual interpretation from 65% to 98%, and of surgical biopsy close to 100%. Therefore, mammography lacks sensitivity, FNA sensitivity varies widely, and surgical biopsy, although accurate, is invasive, time consuming, and costly. The goal of the diagnostic aspect of this research is to develop a relatively objective system.

**Table 1. The description about the attributes of the dataset**

| No  | Attribute                   | Domain                         |
|-----|-----------------------------|--------------------------------|
| 1.  | Sample code number          | Id-number                      |
| 2.  | Clump thickness             | 1-10                           |
| 3.  | Uniformity of cell size     | 1-10                           |
| 4.  | Uniformity of cell shape    | 1-10                           |
| 5.  | Marginal Adhesion           | 1-10                           |
| 6.  | Single Epithelial cell size | 1-10                           |
| 7.  | Bare Nuclei                 | 1-10                           |
| 8.  | Bland Chromatin             | 1-10                           |
| 9.  | Normal Nucleoli             | 1-10                           |
| 10. | Mitoses                     | 1-10                           |
| 11. | Class                       | (2 for benign, 4 for malignant)|

A crucial and important requirement of SVMs is the problem of the optimal feature selection. The optimal selection of features is important in realizing high performance classification. Because SVMs are trained so that the margins are maximized, they are said to be robust for non-optimal features. Moreover, there are 9 features [2 to 10 in Table 1] related to the disease, and not all of these are useful for the task of disease prediction. Therefore some extra selection process has to be carried out so as to narrow down the set of features and get a better prediction rate. Feature selection plays an important role in building classification systems [LLW02]. It not only reduces the dimension of data, but also lowers the computation costs and gains a good classification performance. F-score method is used to select the features

## 1.2. The F-score Method

The F-score is used to measure the discrimination of two sets of real numbers [Ak09]. Given training vectors $x_k$; k = 1; 2; _ _ _; l, and the number of positive and negative instances are $n_+$ and $n_-$ respectively, then the F-score of the *ith* feature is defined as:

$$F_i = \frac{(\overline{x_i}^{(+)} - \overline{x_i})^2 + (\overline{x_i}^{(-)} - \overline{x_i})^2}{\frac{1}{(n_+ - 1)}\sum_{k=1}^{n_+}(x_{k,i}^{(+)} - \overline{x_i}^{(+)})^2 + \frac{1}{(n_- - 1)}\sum_{k=1}^{n_-}(x_{k,i}^{(-)} - \overline{x_i}^{(-)})^2} \quad (1)$$

where $\overline{x}_i$, $\overline{x}_i^{(+)}$ and $\overline{x}_i^{(-)}$ are the average of the i$^{th}$ feature of the whole, positive and negative datasets respectively, and $x_{k,i}^{(+)}$ is the i$^{th}$ feature of the k$^{th}$ positive instance and $x_{k,i}^{(-)}$ i$^{th}$ feature of the k$^{th}$ negative instance. The numerator of the above equation indicates the discrimination between the positive and negative sets and denominator the one within each subset. The larger the F-score is, the more likely this feature is discriminative. This score is hence used as a feature selection criterion.

The procedure is summarized as follows:

1. Initialize the destination subset empty and source subset with all n features
2. Calculate the f-score for each feature as described in equation 1.
3. Add the feature to the destination subset.
4. Go to step 2 until all features in the source subset have been processed.
5. Sort the destination subset features in descending order of the f-score.
6. Train the sample set with SVM leaving the features with least score one by one.
7. Go to step 6 until the sample set contains top 5 features.

## 2. Results

Using gist SVM, one can get the results as in Table 2 and Table 3. From the results, we see that the gist SVM is able to train a classifier with weight and discriminant. Let D be the discriminant value. Then, according to the discriminant value D, we can classify the new data point into the positive class if D > 0, and classify it into the negative class when D < 0.

**Table 2. Results for Training set**

| Example | class | weight | train_classification | train_discriminant |
|---------|-------|--------|----------------------|--------------------|
| 1165926 | 1 | 0 | 1 | 4.226 |
| 1112209 | 1 | 0 | 1 | 3.922 |
| 1116116 | 1 | 0 | 1 | 3.724 |
| 1116998 | 1 | 0.003103 | 1 | 0.9802 |
| 1113483 | 1 | 0.01813 | 1 | 0.8974 |
| 1102573 | 1 | 0.02076 | 1 | 0.8893 |
| 1133041 | -1 | -0.08992 | -1 | -0.5062 |
| 1121919 | -1 | -0.08375 | -1 | -0.5315 |
| 1143978 | -1 | -0.04916 | -1 | -0.7296 |
| 1105524 | -1 | -0 | -1 | -1.2 |
| 1116192 | -1 | -0 | -1 | -1.2 |
| 1147044 | -1 | -0 | -1 | -1.228 |
| 1137156 | -1 | -0 | -1 | -1.274 |

**Table 3. Results for Test set**

| Example | classification | discriminant |
|---------|----------------|--------------|
| 1106829 | 1 | 1.81912 |
| 1112209 | 1 | 1.66274 |
| 1033078 | 1 | 1.66228 |
| 1105257 | 1 | 1.47566 |
| 1054590 | 1 | 0.167441 |
| 1041801 | 1 | 0.00841774 |
| 1048672 | -1 | -0.110634 |
| 1072179 | -1 | -0.154081 |
| 1049815 | -1 | -0.164895 |
| 1050718 | -1 | -0.204936 |

## 3. Explanation for the results

- Example: The name provided for the sample
- Class (training results only): The class membership provided for the sample.
- Weight (training results only): The 'importance' of this example in setting the location of the decision boundary (which is the maximum margin hyper plane). Examples with non-zero weights are support vectors.

- Train_classification (training results only) or classification: The predicted class of this example, or, for training, the location of this example with respect to the decision boundary. In training, if it differs from the Class, a training error is counted.
- Train_discriminant (training results only) or discriminant (test results): How far this example is from the decision boundary. Larger values correspond to greater 'certainty' that the sample belongs in the predicted class.

To have a better classification rate, we need to consider the size of the training set and also the selection of optimal features. In other words, we need to find the optimal size of the training set and the features. In training a SVM, the decision boundaries are determined directly from the training data so that the generalization ability is maximized. Moreover, one needs to solve a quadratic optimization problem with the number of variables equal to the number of training data. We note that determining the optimal size of the training set is important in improving the classification rate. To find the optimal size of the training set, we will try different sizes of training sets and see which one can achieve the highest classification rate. Table 4 gives the classification rate when the number of training-test partitions is 50-50, 60-40, 70-30, 80-20 and 40-60 respectively. From Table 4, one can see that the best classification rate is reached when we choose 50-50 training -test partitions.

**Table 4. Results for the various training-test set partitions**

| Training-Test partitions | Training set accuracy (%) | Test set accuracy (%) |
|---|---|---|
| 50-50 | 100 | 96 |
| 60-40 | 98 | 92.5 |
| 70-30 | 89 | 87.6 |
| 80-20 | 86 | 89.9 |
| 40-60 | 98 | 95 |

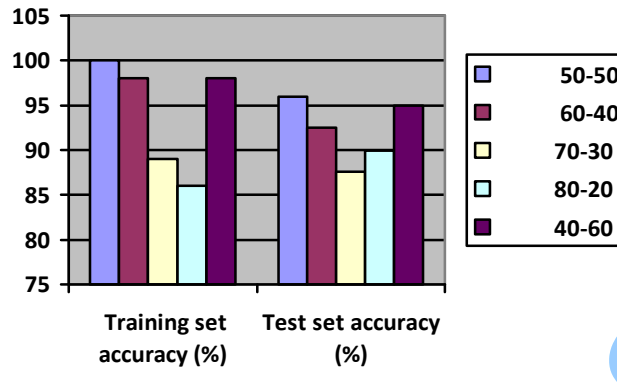

**Figure 2. Bar chart for Table 4**

To investigate potential predictive value of these candidates, we next rank these 9 features based on their scores generated using F-score. Then we train the SVM. Here we used 4 sample sets to train SVM (Set 1 contains datasets with features 1, 2,3,4,5,6,7,9, Set 2 contains datasets with features 1,2,3,4,6,7,9, Set 3 contains datasets with features 1, 2, 3,6,7,9 and Set 4 contains datasets with features 1, 3, 6, 7, 9). The results obtained are presented in Table 5. From the table, we see that the best classification rate is reached when we choose the top five features. 2,4,5,8 considered to be weak features according to the f-score. After removing those features one by one did not make much difference in the accuracy of classification of the samples.

**Table 5. Classification rate for various datasets**

| Sample set | Training set accuracy (%) | Testing set accuracy (%) |
|---|---|---|
| Set 1 | 98.1 | 95.9 |
| Set 2 | 98.6 | 96.1 |
| Set 3 | 97.3 | 96.2 |
| Set 4 | 96.2 | 96.4 |

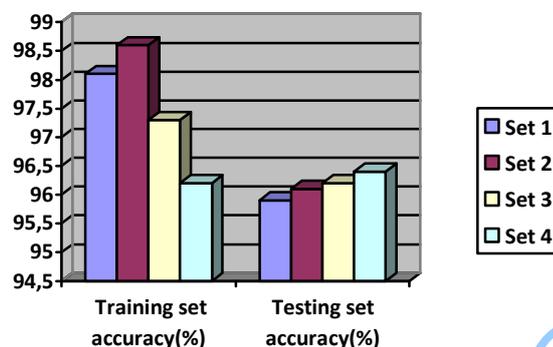

**Figure 3. Bar chart for Table 5**

**Conclusion**

In this study we used SVM based diagnostic model for the diagnosis of breast cancer using Wisconsin dataset. Here we trained the datasets in Gist. In this study we tried to find out the optimum size of the training test partitions and optimum number of features for the dataset using SVM and F-score combination. In future work we proposed to use various kernel functions with SVM for better accuracy and compare the studies with various machine learning tools.